\documentclass{article}
\usepackage{spconf,amsmath,graphicx,hyperref}
\usepackage{amssymb}
\usepackage{algorithm}
\usepackage{algpseudocode}
\usepackage{booktabs}
\usepackage{multirow} 

\usepackage[table]{xcolor}

\title{Sim-MSTNet: sim2real based Multi-task SpatioTemporal Network Traffic Forecasting}
\name{Hui Ma$^{1,\ast}$, Qingzhong Li$^{1}$, Jin Wang$^{1}$, Jie Wu$^{1}$, Shaoyu Dou$^{2}$, Li Feng$^{3}$, Xinjun Pei$^{1}$\thanks{%
This work was supported by the Tianchi Talents - Young Doctor Program (5105250183m),  Science and Technology Program of Xinjiang Uyghur Autonomous Region (2024B03028, 2025B04051), Regional Fund of the National Natural Science Foundation of China (202512120005). For Shaoyu Dou: This work was conducted at Tongji University and is unrelated to the author's current affiliation with Ant Group.
\textsuperscript{*}Corresponding author: Hui Ma, email: \texttt{huima@xju.edu.cn}.}}
\address{$^{1}$Xinjiang Key Laboratory of Intelligent Computing and Smart Applications,\\
School of Software, Xinjiang University, Urumqi, 830046, China\\
$^{2}$Department of Computer Science and Technology, Tongji University, Shanghai,China\\
$^{3}$Bielefeld University of Applied Sciences and Arts, Bielefeld, Germany\\
}

\begin{document}
\ninept
\maketitle
\begin{abstract}
Network traffic forecasting plays a crucial role in intelligent network operations, but existing techniques often perform poorly when faced with limited data. Additionally, multi-task learning methods struggle with task imbalance and negative transfer, especially when modeling various service types. To overcome these challenges, we propose Sim-MSTNet, a multi-task spatiotemporal network traffic forecasting model based on the sim2real approach. Our method leverages a simulator to generate synthetic data, effectively addressing the issue of poor generalization caused by data scarcity. By employing a domain randomization technique, we reduce the distributional gap between synthetic and real data through bi-level optimization of both sample weighting and model training. Moreover, Sim-MSTNet incorporates attention-based mechanisms to selectively share knowledge between tasks and applies dynamic loss weighting to balance task objectives. Extensive experiments on two open-source datasets show that Sim-MSTNet consistently outperforms state-of-the-art baselines, achieving enhanced accuracy and generalization.
\end{abstract}
\section{Introduction}
\label{sec:intro}
With the rapid development of Sixth-Generation (6G) network planning~\cite{11017620}, cellular communication systems are expanding both in scale and complexity, leading to an unprecedented growth in device connectivity and data volume~\cite{10608156}. The increasing number and diversity of connected devices introduce significant heterogeneity and dynamic variations in network topology and traffic patterns~\cite{10443962}. Moreover, the exponential surge in traffic exacerbates network uncertainty, often resulting in bursty loads and anomalous fluctuations~\cite{alwis2021survey}. These challenges make it increasingly difficult to maintain efficient and intelligent network operations, highlighting the need for adaptive management solutions for next-generation communication infrastructure~\cite{10345462}.

Accurate network traffic forecasting is crucial for enabling intelligent network decision-making. Mobile network traffic, however, inherently exhibits complex spatiotemporal correlations, which has driven the development of advanced deep learning models capable of capturing both temporal dynamics and spatial dependencies. Despite their effectiveness, the performance of these models is heavily dependent on large-scale, high-quality data~\cite{jiang2022trendgcn}. In scenarios with limited data, such as newly deployed base stations, remote areas, or emergency communication networks, the severe lack of training data significantly hampers model generalization, resulting in considerable degradation in prediction accuracy~\cite{mo2022trafficflowgan}.

To address these limitations, previous research has primarily focused on two categories of solutions. The first category involves data generation techniques, such as Generative Adversarial Networks (GANs)~\cite{10446144} and diffusion models~\cite{lu2025diffusion, chai2024fomo}. However, these models often struggle to capture the complex spatiotemporal structures of real-world traffic, leading to a significant distribution gap between generated synthetic data and authentic data. The second category is the sim2real technique~\cite{10742102}, which uses physical simulators to generate large amounts of training data. However, it faces the "reality gap," as simulation environments cannot accurately replicate the complexity and dynamics of real-world wireless propagation. As a result, models trained on simulated data often fail to generalize effectively to real-world scenarios~\cite{muruganandham2025smart, da2025survey}.

Another core challenge lies in effectively addressing conflicts and imbalance in Multi-Task Learning (MTL). In network traffic forecasting, tasks are often heterogeneous (e.g., Call, SMS, and Net traffic), and traditional single-task learning models fail to exploit the potential correlations among them. MTL, by contrast, improves model generalization and efficiency through learning shared representations~\cite{wang2023grec}. However, most existing traffic MTL approaches rely on hard parameter-sharing mechanisms~\cite{ma2018modeling}, which are particularly vulnerable to task conflicts: Certain tasks may dominate the optimization process while others are suppressed, ultimately causing negative transfer and degrading overall performance.

To effectively address the aforementioned challenges, this paper proposes Sim-MSTNet, a unified framework that simultaneously tackles both data scarcity and task conflict issues. \textit{Specifically}, we model wireless communication environments and generate extensive simulated data under multiple configurations. To narrow the reality gap, domain randomization is employed to train the model on reweighted simulated samples. \textit{Subsequently}, the joint optimization of model parameters and sample weights is formulated as a bi-level optimization problem. \textit{Then}, we develop a cutting-plane-based optimization algorithm to enhance the generalization performance of network traffic time series forecasting models. \textit{Besides}, to accurately model the complex spatiotemporal dependencies in network traffic, we incorporate a dynamic multi-task network, which leverages an attention mechanism to achieve soft parameter sharing and information replenishment among different tasks (e.g., Call, SMS, and Net). \textit{Furthermore}, a dynamic loss weighting strategy is adopted to adaptively balance tasks, which helps mitigate task imbalance and reduce negative transfer.

The main contributions of this paper are summarized as follows. \textit{Firstly}, we propose a deep learning model with sim2real technique and leverage the domain randomization based method to solve the "reality gap" issue. In fact, we are the first to employ a sim2real technique with a domain randomization strategy for cellular network traffic prediction. \textit{Besides}, we introduce a multi-task spatiotemporal approach with a dynamic loss weighting strategy. It is the first work to explore dynamic loss weighting strategy specifically for cellular network traffic prediction. \textit{Furthermore}, extensive experiments on two open-sourced datasets demonstrated that Sim-MSTNet outperforms state-of-the-art baselines.

\begin{figure*}[t]
  \centering
   \includegraphics[width=0.8\linewidth,clip,trim=8mm 70mm 8mm 25mm]{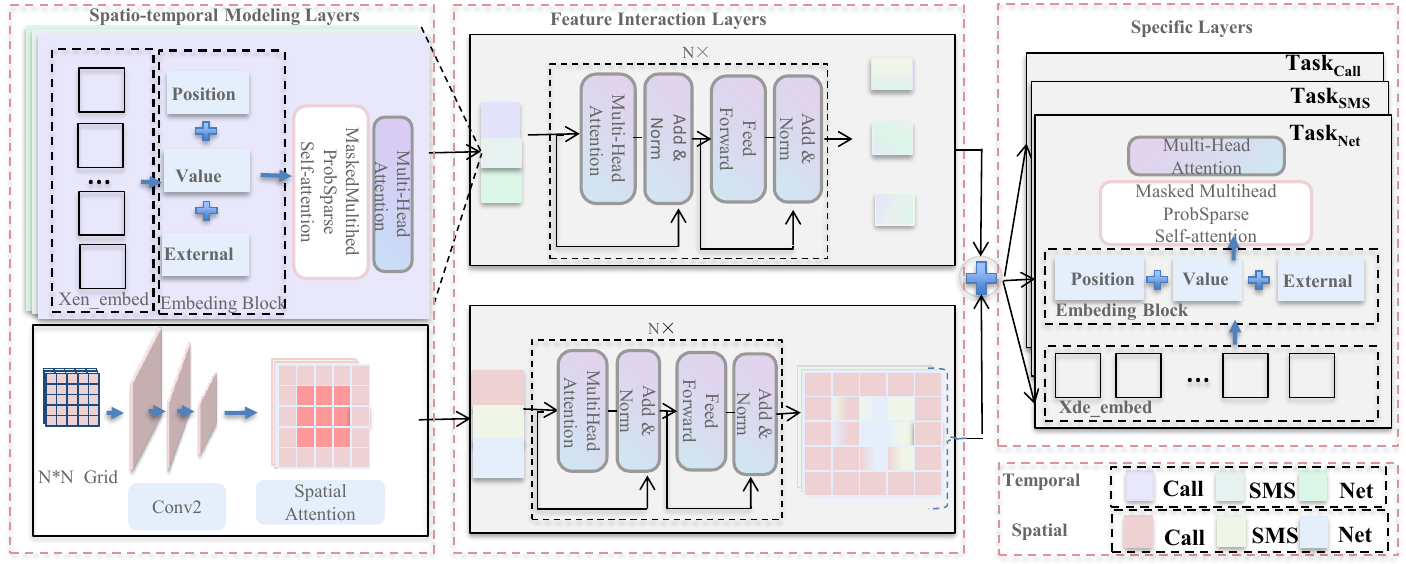}
   \caption{Structure of Sim-MSTNet model.
   }
   \label{fig:framework}
\end{figure*}
\section{METHOD}
\label{sec:methodology}
\subsection{Problem Formulation}

We formulate sim2realtransfer as a bilevel optimization problem and employ domain randomization through sample reweighting,

\begin{equation}
\footnotesize
\begin{aligned}
\min_{\boldsymbol{w}} \quad & G(\boldsymbol{w}, \boldsymbol{\phi}) = \frac{1}{|\mathcal{D}_v|} \sum_{(\boldsymbol{x}_i, y_i) \in \mathcal{D}_v} \mathcal{L}(\boldsymbol{x}_i, y_i, \boldsymbol{\phi}^*(\boldsymbol{w})), \\
\text{s.t.} \quad & \boldsymbol{\phi}^*(\boldsymbol{w}) = \arg\min_{\boldsymbol{\phi}} \frac{1}{|\mathcal{D}_s|} \sum_{(\boldsymbol{x}_i, y_i) \in \mathcal{D}_s} \sigma(w_j) \mathcal{L}(\boldsymbol{x}_i, y_i, \boldsymbol{\phi}),
\end{aligned}
\label{eq:reweighting}
\end{equation}
where $\boldsymbol{w}$ are sample weights, $\boldsymbol{\phi}$ are model parameters, $\mathcal{D}_v$ is the real validation dataset, $\mathcal{D}_s$ is the simulated dataset, and $\sigma(\cdot)$ is the sigmoid function.

Problem~\eqref{eq:reweighting} is extremely difficult to solve due to its non-smooth and non-convex nature. To address this challenge, we employ a K-step gradient descent approximation to replace the intractable lower-level optimization. Specifically, we approximate $\boldsymbol{\phi}^*(\boldsymbol{w})$ through the following iterative process,

\begin{equation}
\boldsymbol{\phi}_{k+1} = \boldsymbol{\phi}_k - \eta \nabla_{\boldsymbol{\phi}} g(\boldsymbol{w}, \boldsymbol{\phi}_k), \quad k = 0, 1, \ldots, K-1,
\label{eq:k_step_gd}
\end{equation}
where $g(\boldsymbol{w}, \boldsymbol{\phi}) = \frac{1}{|\mathcal{D}_s|} \sum_{(\boldsymbol{x}_i, y_i) \in \mathcal{D}_s} \sigma(w_j) \mathcal{L}(\boldsymbol{x}_i, y_i, \boldsymbol{\phi})$ is the weighted loss on simulated data. The approximate solution is then given by:

\begin{equation}
\psi(\boldsymbol{w}) = \boldsymbol{\phi}_0 - \eta \sum_{k=0}^{K-1} \nabla_{\boldsymbol{\phi}} g(\boldsymbol{w}, \boldsymbol{\phi}_k).
\label{eq:psi_approximation}
\end{equation}

This approximation transforms the bilevel optimization into a single-level constrained problem where the constraint becomes $h(\boldsymbol{w}, \boldsymbol{\phi}) = \|\boldsymbol{\phi} - \psi(\boldsymbol{w})\|_1 \leq \epsilon$, making the problem computationally tractable while preserving the essential structure of the original formulation.
\subsection{Cutting-Plane-based Optimization Algorithm}

To solve the constrained optimization problem efficiently, we develop a two-phase algorithm that dynamically manages the polyhedral approximation. The algorithm alternates between updating optimization variables and refining the polyhedron by adding/removing cutting planes.

The algorithm operates in two phases with different update strategies,

\textbf{Phase 1} ($t < T_1$): Dynamic polyhedron updates
\begin{align}
\boldsymbol{w}^{t+1} &= \boldsymbol{w}^t - \eta_{\boldsymbol{w}} \nabla_{\boldsymbol{w}} L_q, \quad \boldsymbol{\phi}^{t+1} = \boldsymbol{\phi}^t - \eta_{\boldsymbol{\phi}} \nabla_{\boldsymbol{\phi}} L_q, \label{eq:update_w}\\
\mu_l^{t+1} &= \mu_l^t + \eta_{\mu_l} \nabla_{\mu_l} L_q. \label{eq:update_mu}
\end{align}

Every $\delta$ iterations: remove inactive planes ($\mu_l = 0$) and add new planes when $h(\boldsymbol{w}^{t+1}, \boldsymbol{\phi}^{t+1}) > \epsilon$. New cutting plane coefficients are derived from the gradient of $h$,

\begin{align}
\boldsymbol{a}_l &= \frac{\partial h(\boldsymbol{w}^{t+1}, \boldsymbol{\phi}^{t+1})}{\partial \boldsymbol{w}}, \quad \boldsymbol{b}_l = \frac{\partial h(\boldsymbol{w}^{t+1}, \boldsymbol{\phi}^{t+1})}{\partial \boldsymbol{\phi}}, \\
c_l &= h(\boldsymbol{w}^{t+1}, \boldsymbol{\phi}^{t+1}) - \left[\frac{\partial h}{\partial \boldsymbol{w}}, \frac{\partial h}{\partial \boldsymbol{\phi}}\right]^{\top} \left[\boldsymbol{w}^{t+1}, \boldsymbol{\phi}^{t+1}\right]^{\top}.
\end{align}

\textbf{Phase 2} ($t \geq T_1$): Fixed polyhedron with penalty method
\begin{align}
\boldsymbol{w}^{t+1} &= \boldsymbol{w}^t - \eta_{\boldsymbol{w}} \nabla_{\boldsymbol{w}} \hat{L}_q, \quad \boldsymbol{\phi}^{t+1} = \boldsymbol{\phi}^t - \eta_{\boldsymbol{\phi}} \nabla_{\boldsymbol{\phi}} \hat{L}_q, \label{eq:penalty_w}
\end{align}
where $\hat{L}_q = G(\boldsymbol{w}, \boldsymbol{\phi}) + \sum_{l} \mu_l (\max\{0, \boldsymbol{a}_l^{\top} \boldsymbol{w} + \boldsymbol{b}_l^{\top} \boldsymbol{\phi} + c_l\})^2$.

The two-phase design balances exploration (Phase 1) and exploitation (Phase 2), ensuring both adaptive constraint refinement and stable convergence.

\begin{algorithm}[t]
\caption{Cutting-Plane Sample Reweighting}
\label{alg:cutting_plane}
\begin{algorithmic}[1]
\State \textbf{Input}: $\boldsymbol{w}^0, \boldsymbol{\phi}^0, \mathcal{P}^0, \boldsymbol{\mu}^0$
\State $t = 0$
\While{not converged}
\If{$t < T_1$}
\State Update $\boldsymbol{w}^{t+1}, \boldsymbol{\phi}^{t+1}, \boldsymbol{\mu}^{t+1}$ via Eqs.~\eqref{eq:update_w}-\eqref{eq:update_mu}
\If{$(t+1) \bmod \delta == 0$}
\State Update polyhedron $\mathcal{P}^{t+1}$ (remove inactive, add new planes)
\EndIf
\Else
\State Update $\boldsymbol{w}^{t+1}, \boldsymbol{\phi}^{t+1}$ via Eqs.~\eqref{eq:penalty_w}
\EndIf
\State $t = t + 1$
\EndWhile
\State \textbf{Output}: $\boldsymbol{w}, \boldsymbol{\phi}, \mathcal{P}, \boldsymbol{\mu}$
\end{algorithmic}
\end{algorithm}

\subsection{Sim-MSTNet Architecture}
Figure \ref{fig:framework} presents the Sim-MSTNet framework for multi-task cellular traffic prediction. The architecture comprises three key components: (1) spatiotemporal modeling layer for feature extraction, (2) feature interaction layer with task-level attention for cross-task information sharing, and (3) task-specific decoding layer with dynamic loss weighting to mitigate negative transfer.

\subsubsection{Spatiotemporal Feature Extraction}
The spatiotemporal modeling layer captures both temporal and spatial dependencies. For temporal modeling, (*) represents Call, SMS, and Net tasks, at each time step $t$, the model input consists of raw traffic sequence $x_t^{\text{Call}}$ and temporal markers. We employ embedding mechanisms combining value, external, and positional encodings,
\begin{equation}
	R_{t,\text{enc}}^{\text{*}} = V_{t,\text{enc}}^{\text{*}} + E_{t,\text{enc}}^{\text{*}} + P_t^{\text{*}},
\end{equation}
where $V$, $E$, and $P$ represent value, external temporal, and positional embeddings respectively.
 The complete sequence $R_{\text{enc}}^{\text{Call}}$ is then fed into an attention-based encoder to extract contextual representations $H_{\text{enc}}^{\text{Call}}$ in the temporal dimension.

For spatial modeling, input data $X^{\text{task,grid}}$ is processed through CNN followed by multi-head spatial attention,
\begin{equation}
	H_{s}^{\text{*}} = \text{Linear}(\text{MultiHeadAtt}_{\text{spatial}}(f_{\text{CNN}}(X^{\text{*,grid}}))),
\end{equation}
where $*$ represents the Call, SMS, and Net tasks.

\subsubsection{Multi-Task Feature Interaction}
To exploit collaborative relationships between Call, SMS, and Net tasks, we design a feature interaction module that constructs task-to-task attention mechanisms. The spatial and temporal features from all tasks are concatenated,
\begin{equation}
	H_s^{\text{cat}} = [H_s^{\text{Call}}; H_s^{\text{SMS}}; H_s^{\text{Net}}], \quad H_t^{\text{cat}} = [H_{\text{enc}}^{\text{Call}};H_{\text{enc}}^{\text{SMS}};H_{\text{enc}}^{\text{Net}} ].
\end{equation}

Cross-task dependencies are modeled through multi-head attention, where spatial and temporal features are processed separately to capture task interactions. The attention mechanism outputs enhanced representations $\hat{H}_s$ and $\hat{H}_t$ that encode cross-task dependencies.

After residual connections and layer normalization, spatial and temporal representations are fused as $Z = [\hat{H}_s; \hat{H}_t]$, then fed into task-specific MLPs to generate interaction-aware representations $\tilde{H}^{\text{Call}}$, $\tilde{H}^{\text{SMS}}$, $\tilde{H}^{\text{Net}}$.
\subsubsection{Task-Specific Decoding Layer}
In the decoder module, the raw decoder input is first embedded using the same scheme as the encoder (i.e., value, external-feature, and positional encodings) to obtain the task-specific representation $R_{\mathrm{dec}}^{*}$. The decoder then employs masked multi-head self-attention and ProbSparse cross-attention for efficient prediction,
\begin{equation}
    Y^{*} = \mathrm{Decoder}_{*}(R_{\mathrm{dec}}^{*}, \tilde{H}^{*}).
\end{equation}

\subsubsection{Dynamic Loss Weighting Strategy}
To address task imbalance and negative transfer, we introduce a dynamic weighting mechanism that adaptively adjusts task contributions during training. The weight update rule for task $i$ at iteration $k$ is,
\begin{equation}
	w_*^{(k+1)} = (1 - \alpha) \cdot w_*^{(k)} + \alpha \cdot \left( \frac{\mathcal{L}_*^{(k)}}{\sum_{*} \mathcal{L}_*^{(k)}} \right),
\end{equation}
where $\alpha$ is the smoothing factor and $\mathcal{L}_i^{(k)}$ is the loss for task $i$.

Then, the total loss can be obtained, 
\begin{equation}
	\mathcal{L}_{\text{total}}^{(k)} = \sum_{* \in \{\text{Call, SMS, Net}\}} w_*^{(k)} \cdot \mathcal{L}_*^{(k)}.
\end{equation}

\subsection{Dataset Description}
We evaluate our method on three datasets: (1) \textbf{Simulation Data}: 30,638 synthetic samples and 982 real samples from cellular networks using Wireless InSite simulator with domain randomization; (2) \textbf{Milano}: Cellular traffic data from Milan, Italy (Nov 2013 - Jan 2014) with 10,000 grid cells covering SMS, Call, and Internet activities; (3) \textbf{Trento}: Telecom Italia data from Trento Province (Nov-Dec 2013) with 6,575 grid cells recording similar communication activities.


\section{Experimental Results}\label{sec5}

\subsection{Single-Task Comparison}

Table~\ref{tab:grouped_comparison} compares STNet-Sim2RealBO with eight representative single-task baseline models, including traditional temporal methods (LSTM, Transformer), enhanced Transformers (Informer, iTransformer), and spatiotemporal modeling methods (ConvLSTM, STGCN, DCRNN, STAEformer). All models are trained on the Net task and evaluated on Milano and Trento datasets using MAE and RMSE metrics.

Traditional models such as LSTM and Transformer focus on temporal dependencies while neglecting spatial structures, resulting in limited predictive accuracy. Enhanced Transformers like Informer and iTransformer improve long-range dependency modeling but remain limited to temporal modeling. Spatiotemporal methods provide notable improvements but exhibit important limitations: STGCN and DCRNN rely on predefined static graphs, ConvLSTM's limited receptive field hinders global spatial modeling, and STAEformer's performance degrades under data-scarce conditions.

STNet-Sim2RealBO integrates CNNs, spatial attention modules, and Informer-based decoders while introducing a cutting-plane bi-level optimization strategy. Unlike prior models, STNet-Sim2RealBO employs simulated data as the primary training source and dynamically optimizes sample weights to mitigate sim2real distribution bias. Leveraging spatial modeling, long-term dependency learning, and adaptive sample quality control, STNet-Sim2RealBO consistently achieves superior results across both datasets.

\subsection{Multi-Task Prediction Results}
\begin{table}[t]
	\small
	\centering
	\caption{Single-task performance}
	\label{tab:grouped_comparison}
	\begin{tabular}{l|cc|cc}
		\toprule
		\textbf{Models} 
		& \multicolumn{2}{c|}{\textbf{Milano}} 
		& \multicolumn{2}{c}{\textbf{Trento}} \\
		& MAE & RMSE & MAE & RMSE \\
		\midrule
		LSTM          & 6.49  & 10.75 & 16.90  & 23.63 \\
		Transformer   & 6.15  & 10.77 & 14.41  & 20.90 \\
		Informer~\cite{Zhou2021Informer}      & 5.18  & 9.37  & 7.10   & 10.56 \\
		iTransformer~\cite{liu2023itransformer}  & 5.14  & 9.52  & 5.71   & 8.71  \\
		\midrule
		STGCN         & 5.22  & 9.64  & 7.72   & 11.52 \\
		DCRNN         & 5.05  & 8.97  & 5.96   & 9.16  \\
		STAEformer~\cite{liu2023staeformer}    & 5.25  & 8.94  & 6.08   & 9.76  \\
		\midrule
		\textbf{STNet-Sim2RealBO} & \textbf{3.96} & \textbf{7.55} & \textbf{5.63} & \textbf{8.40} \\
		\bottomrule
	\end{tabular}
\end{table}
Table~\ref{tab:ieee_multitask} shows multi-task performance comparison. Our Sim-MSTNet consistently outperforms MTTC, AST-MTL, and CSLSL across all tasks and datasets. It integrates spatial-aware CNN modules with task-specific temporal decoders, enabling effective extraction of local spatial features while capturing long-term temporal dependencies via Informer-based decoders.

 Sim-MSTNet introduces dynamic interaction fusion modules that adaptively capture shared knowledge across tasks while preserving task-specific features, thereby reducing negative transfer. The exponential-smoothing-based dynamic loss weighting mechanism automatically emphasizes harder-to-learn tasks, improving training stability. These innovations collectively enable Sim-MSTNet to achieve the lowest MAE and RMSE across all tasks and datasets, demonstrating its effectiveness in handling diverse multi-task scenarios.

\begin{figure}[t]
  \centering
   \includegraphics[width=0.6\linewidth]{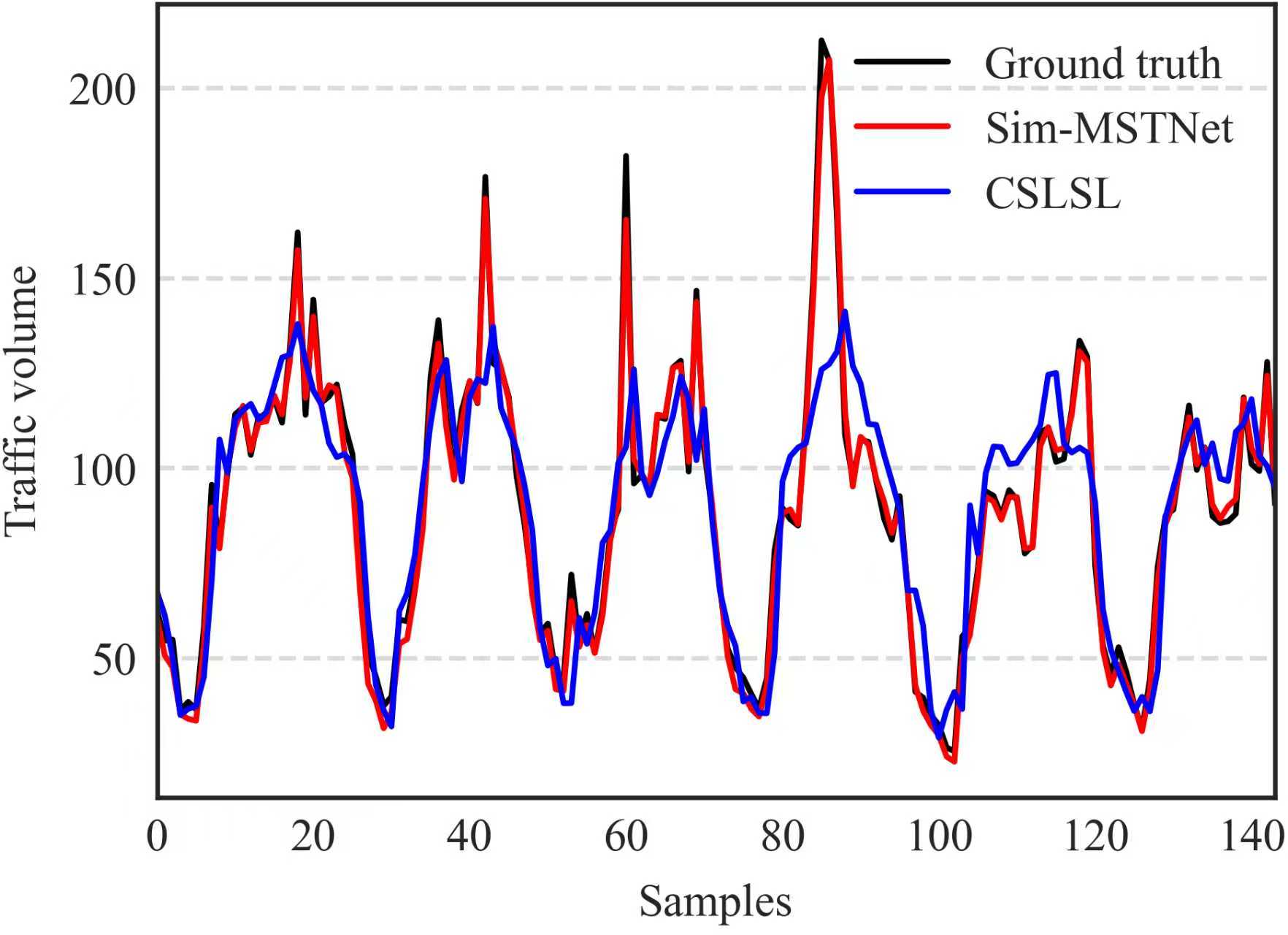}
   \vspace{-6pt}
   \caption{Multi-task prediction visualization results on the Milano dataset.}
   \label{fig:vis_multitask}
\end{figure}
Figure~\ref{fig:vis_multitask} presents multi-task prediction visualization on the Milano dataset. The visualization compares ground truth (black line), the proposed Sim-MSTNet model (red line), and the second-best multi-task baseline model CSLSL (blue line). Sim-MSTNet accurately fits periodic patterns in Milano and tracks high-frequency variations without response delays, consistently exceeding the CSLSL baseline across all scenarios.
\begin{table}[t]
\centering
\caption{Multi-task prediction performance}
\label{tab:ieee_multitask}
\small
\setlength{\tabcolsep}{2.5pt}
\begin{tabular}{l|l|ccc|ccc}
    \toprule
    \textbf{Models} & \textbf{Metric} 
    & \multicolumn{3}{c|}{\textbf{Milano}} 
    & \multicolumn{3}{c}{\textbf{Trento}} \\
    & & Call & SMS & Net & Call & SMS & Net \\
    \midrule
    \multirow{2}{*}{MTTC~\cite{rezaei2020mttc}} 
    & MAE  & 0.96 & 1.73 & 5.15 & 1.25 & 3.61 & 7.46 \\
    & RMSE & 1.65 & 3.83 & 9.73 & 2.18 & 7.25 & 11.62 \\
    \midrule
    \multirow{2}{*}{AST-MTL~\cite{yang2023astmtl}} 
    & MAE  & 1.08 & 1.81 & 5.24 & 1.31 & 4.06 & 7.19 \\
    & RMSE & 1.90 & 4.08 & 9.86 & 2.24 & 8.21 & 10.98 \\
    \midrule
    \multirow{2}{*}{CSLSL~\cite{huang2024human}} 
    & MAE  & 0.94 & 1.54 & 2.51 & 1.50 & 3.41 & 7.00 \\
    & RMSE & 1.53 & 3.28 & 4.51 & 2.69 & 7.10 & 11.10 \\
    \midrule
    \multirow{2}{*}{\textbf{Sim-MSTNet}} 
    & MAE  & \textbf{0.29} & \textbf{0.41} & \textbf{1.40} & \textbf{0.29} & \textbf{0.75} & \textbf{1.89} \\
    & RMSE & \textbf{0.40} & \textbf{0.80} & \textbf{2.09} & \textbf{0.47} & \textbf{1.62} & \textbf{3.33} \\
    \bottomrule
\end{tabular}
\end{table}
\subsection{Ablation Study}
\begin{table}[t]
\centering
\caption{Ablation study on Milano and Trento datasets}
\label{tab:ablation_combined}
\small
\setlength{\tabcolsep}{1.0pt}
\begin{tabular}{l|l|ccc|ccc}
    \toprule
    \textbf{Models} & \textbf{Metric} 
    & \multicolumn{3}{c|}{\textbf{Milano}} 
    & \multicolumn{3}{c}{\textbf{Trento}} \\
    \cmidrule(lr){3-5} \cmidrule(lr){6-8}
    & & Call & SMS & Net & Call & SMS & Net \\
    \midrule
    \multirow{2}{*}{Sim-MSTNet/oInter} 
    & MAE  & 0.65 & 0.99 & 2.85 & 0.73 & 3.54 & 6.36 \\
    & RMSE & 0.93 & 2.09 & 3.75 & 1.07 & 7.06 & 9.32 \\
    \midrule
    \multirow{2}{*}{Sim-MSTNet/oSpatial} 
    & MAE  & 0.55 & 0.78 & 1.63 & 0.36 & 1.06 & 2.07 \\
    & RMSE & 0.99 & 2.06 & 2.93 & 0.54 & 2.36 & 3.68 \\
    \midrule
    \multirow{2}{*}{Sim-MSTNet(Average)} 
    & MAE  & 0.55 & 0.78 & 1.93 & 0.38 & 1.31 & 3.06 \\
    & RMSE & 1.01 & 2.20 & 3.44 & 0.59 & 2.68 & 4.76 \\
    \midrule
    \multirow{2}{*}{\textbf{Sim-MSTNet}} 
    & MAE  & \textbf{0.29} & \textbf{0.41} & \textbf{1.40} & \textbf{0.29} & \textbf{0.75} & \textbf{1.89} \\
    & RMSE & \textbf{0.40} & \textbf{0.80} & \textbf{2.09} & \textbf{0.47} & \textbf{1.62} & \textbf{3.33} \\
    \bottomrule
\end{tabular}
\end{table}
Table~\ref{tab:ablation_combined} validates each component's contribution. Removing spatial modeling significantly degrades performance because cellular traffic exhibits strong spatial correlations that require explicit modeling to capture inter-cell dependencies. Removing task interaction also leads to substantial degradation, as the shared representations between call and SMS tasks contain complementary information that enhances prediction accuracy when properly fused. Dynamic loss weighting outperforms simple averaging by adaptively balancing task competition during training, preventing dominant tasks from overshadowing others and ensuring stable convergence across all objectives.

\section{CONCLUSION}
This paper addresses task imbalance and negative transfer in multi-task cellular traffic prediction by proposing Sim-MSTNet, a dynamic weighted spatiotemporal prediction model. We introduce a bi-level optimization algorithm using cutting-plane methods that incorporates real samples and dynamically reweights large-scale simulated data. Sim-MSTNet integrates CNNs with attention mechanisms for spatiotemporal modeling and employs dynamic loss weighting to mitigate task interference. Experimental results on two open-sourced datasets demonstrate consistent improvements over state-of-the-art baselines.

\bibliographystyle{IEEEtran}
\bibliography{refs}

\end{document}